\documentclass[conference]{IEEEtran}
\usepackage[numbers, sort, comma, square]{natbib}

\usepackage[utf8]{inputenc} 
\usepackage[T1]{fontenc}    
\usepackage{hyperref}       
\usepackage{url}            
\usepackage{booktabs}       
\usepackage{rotating}
\usepackage{makecell}
\usepackage{multirow}
\usepackage{siunitx}
\usepackage{amsfonts}       
\usepackage{nicefrac}       
\usepackage{microtype}      
\usepackage{siunitx,etoolbox}
\usepackage{bm}
\usepackage{amsmath}
\usepackage{subcaption}
\usepackage{amsthm}
\usepackage[ruled,vlined, linesnumbered]{algorithm2e}

\usepackage{todonotes}
\usepackage{xspace}
\usepackage{amsthm}
\newtheorem{RQ}{RQ}
\newtheorem*{PS}{Problem Statement}
\newcommand{\mpara}[1]{\medskip\noindent{\bf #1}}
\newcommand{\graphsage}{\textsc{Sage}\xspace}
\newcommand{\gcn}{\textsc{Gcn}\xspace}
\newcommand{\gat}{\textsc{Gat}\xspace}
\newcommand{\sgc}{\textsc{Sgc}\xspace}
\newcommand{\cora}{\textsc{Cora}\xspace}
\newcommand{\cseer}{\textsc{CiteSeer}\xspace}
\newcommand{\pubmed}{\textsc{PubMed}\xspace}
\newcommand{\flickr}{\textsc{Flickr}\xspace}
\newcommand{\reddit}{\textsc{Reddit}\xspace}

\newcommand{\vanpd}{\textsc{VanPd}\xspace}
\newcommand{\nsd}{\textsc{NsD}\xspace}
\newcommand{\lbp}{\textsc{LbP}\xspace}

\newtheoremstyle{mystyle}
  {}
  {}
  {\itshape}
  {}
  {\bfseries}
  {.}
  { }
  {\thmname{#1}\thmnumber{ #2}\thmnote{ (#3)}}

\theoremstyle{mystyle}

\newtheorem{theorem}{Theorem}
\newtheorem{definition}[theorem]{Definition}

\usepackage[switch]{lineno}

\begin{document}

\title{Membership Inference Attack on Graph Neural Networks}


\author{
{\rm Iyiola E.\ Olatunji}\\
L3S Research Center,\\
Hannover, Germany. \\
iyiola@l3s.de
\and
{\rm Wolfgang Nejdl}\\
L3S Research Center,\\
Hannover, Germany. \\
nejdl@l3s.de
\and
{\rm Megha Khosla}\\
L3S Research Center,\\
Hannover, Germany. \\
khosla@l3s.de
} 



\maketitle

\begin{abstract}
Graph Neural Networks (GNNs), which generalize traditional deep neural networks on graph data, have achieved state-of-the-art performance on several graph analytical tasks. 
We focus on how trained GNN models could leak information about the \emph{member} nodes that they were trained on.
We introduce two realistic settings for performing a membership inference (MI) attack on GNNs. While choosing the simplest possible attack model that utilizes the posteriors of the trained model (black-box access), we thoroughly analyze the properties of GNNs and the datasets which dictate the differences in their robustness towards MI attack.
While in traditional machine learning models, overfitting is considered the main cause of such leakage, we show that in GNNs the additional structural information is the major contributing factor. 
We support our findings by extensive experiments on four representative GNN models. To prevent MI attacks on GNN, we propose two effective defenses that significantly decreases the attacker's inference by up to 60\% without degradation to the target model's performance.
Our code is available at \url{https://github.com/iyempissy/rebMIGraph}.
\end{abstract}

\section{Introduction}

Graph neural networks (GNNs) have gained substantial attention from academia and industry in the past few years with high-impact applications ranging from the analysis of social networks, recommender systems to biological networks. One of the most popular tasks is that of \emph{node classification} in which the goal is to predict the unknown node labels. These models differ from the traditional machine learning (ML) models, in that they use additional relational information among the node instances to make predictions. In fact, the graph convolution-based model \cite{kipf2017semi} which is the most popular class of GNNs embeds graph structure into the model itself by computing representation of a node via recursive aggregation and transformation of feature representations of its neighbors. 
We take the first step in exposing the vulnerability of such models to \emph{membership inference} (MI) attacks. 
In particular, we ask \emph{whether the trained GNN model can be used to identify the input instances (nodes) that it was trained on.} 


  

To motivate the importance of the problem for graphs, 
suppose a researcher has a list of patients infected with COVID19. The researcher is interested in understanding the various factors contributing to the infection. To account for the factors such as their social activity, she might want to utilize knowledge of friendship/social connection known among the patients. She then trains a GNN model on the graph induced on the nodes of interest and  uses the trained node representations as additional input for her disease analysis models. A successful MI attack on the trained model would reveal the list of infected persons even though the model might have not used any disease-related sensitive information.

The goal of MI attack \cite{Shokri17,salem2020updates} is to distinguish between the \emph{target} model's behavior for the inputs it encountered during training from the ones which it did not. The inputs to the attack model are the class probabilities (posteriors) or the confidence values output by the target model for the corresponding data point. Thus, the attacker or adversary only requires \emph{black-box} access to the model where she can query the model on her desired data record and obtain the model predictions (output class probabilities).

While membership inference  has been well studied in the context of traditional ML models \cite{Shokri17, salem2018ml} like convolution neural networks (CNNs) and multilayer perceptron (MLP), GNNs has so far escaped attention. 
Much of the success of MI attacks in traditional ML models has been attributed to the model's tendency to overfit or memorize the dataset \cite{zhang2016understanding}. 
 Overfitting leads to the assignment of high confidence scores to data records seen during training as compared to new data it encountered during testing, making it possible to distinguish between the instance types from the prediction scores. 
We ask \emph{if overfitting in GNNs is also the main contributing factor for successful membership inference.}
We discover that even if a GNN model generalizes well to unseen data, it can still be highly prone to MI risks.
The \emph{encoding of the graph structure into the model} is what makes a GNN powerful 
but it is exactly this property that makes it much more vulnerable to privacy attacks. Therefore, unlike other models, reducing overfitting might not alone lead to higher robustness against privacy risks. 

While we showed that all GNN models are vulnerable to MI attack, we observed differences in attack success rate. We explain these differences in terms of differing dataset and model properties using insights from our large scale experimental analysis. We further develop defense mechanisms based on \emph{output perturbation} and \emph{query neighborhood perturbation} strategies. Our empirical results show that our defenses can effectively defend against MI attacks on GNNs by reducing the attacker's inference by over 60\% with negligible loss in the target model's inference.

To summarize, our key contributions are as follows.
\begin{enumerate}
\item We introduce two realistic settings for carrying out MI attack on GNNs.
\item We perform an extensive experimental study to expose the risks of privacy leakage in GNN models. We further attribute the differences between the model's robustness towards MI attack to the dataset properties and the model architecture.  

\item Contrary to popular belief, we show that for GNNs, lack of overfitting does not guarantee robustness towards privacy attacks. 
\item We propose two defense mechanisms (based on output and query neighborhood perturbation) against MI attacks in GNNs that significantly degrade attack performance without compromising the target model's utility. 

\end{enumerate}

\section{Background and Related Works}

\subsection{Graph Neural Networks}
\label{sec:gnns}
Graph Neural Networks popularized by graph convolutional networks (GCNs) and their variants, 
generalize the convolution operation for irregular graph data. 
These methods encode graph structure directly into the model.
In particular, the node representation is computed by recursive aggregation and transformation of feature representations of its neighbors.

Let $\boldsymbol{x}_{i}^{(\ell)}$ denote the feature representation of node $i$ at layer $\ell$ and ${\mathcal{N}}_{i}$ denotes the set of its 1-hop neighbors.
Formally, the $\ell$-th layer of a general graph convolutional operation can then be described as
\begin{align}
    \boldsymbol{z}_{i}^{(\ell)}=&\operatorname{AGGREGATION}^{(\ell)}\left(\left\{\boldsymbol{x}_{i}^{(\ell-1)},\left\{\boldsymbol{x}_{j}^{(\ell-1)} \mid j \in{\mathcal{N}}_{i}\right\}\right\}\right) \\
    \boldsymbol{x}_{i}^{(\ell)}= &\operatorname{TRANSFORMATION} ^{(\ell)}\left(\boldsymbol{z}_{i}^{(\ell)}\right)
\end{align}
Finally, a softmax layer is applied to the node representations at the last layer (say $L$) for the final prediction of node classes,
\begin{align}
\label{eq:pred}
     \boldsymbol{y}_i\leftarrow \operatorname{softmax}(\boldsymbol{z}_{i}^{(L)}\mathbf{W}),
\end{align}
    where $ \boldsymbol{y}_i \in \mathbb{R}^{c}$, $c$ is the number of classes and $\mathbf{W}$ is a learnable weight matrix. Each element $ \boldsymbol{y}_i(j)$ corresponds to the (predicted) probability (or posterior) that node $i$ is assigned to class $j$. 

We focus on four representative models of this family which differ either on one of the above two steps of aggregation and transformation. In the following, we briefly describe these models and their differences.

\mpara{Graph Convolutional Network (\gcn)} \cite{kipf2017semi}. Let $d_i$ denote the degree of node $i$. The aggregation operation in \gcn is then given as 
\begin{equation}\label{eq:conv}
\boldsymbol{z}_{i}^{(\ell)} \leftarrow \sum_{j\in \mathcal{N}(i)\cup i} \frac{1}{\sqrt{d_{i}d_{j}}} \boldsymbol{x}_{j}^{(\ell-1)}.
\end{equation}
\gcn performs a non-linear transformation over the aggregated features to compute the representation at layer $\ell$.
\begin{equation}\label{eq:transgcn}
\boldsymbol{x}_{i}^{(\ell)} \leftarrow \operatorname{ReLU}\left(\boldsymbol{z}_{j}^{(\ell-1)} \mathbf{W}^{(\ell)}\right).
\end{equation}

\mpara{Simplifying Graph Convolutional Networks (\sgc)} \cite{pmlr-v97-wu19e}.
 The authors in \cite{pmlr-v97-wu19e} argue that the non-linear activation function in \gcn is not critical for the node classification task and completely skips the non-linear transformation step. In particular, in an $L$ layer \sgc model, $L$ aggregation steps are applied as given by \eqref{eq:conv} followed by final prediction (as in \eqref{eq:pred}).

\mpara{Graph Attention Networks (\gat)} \cite{velickovic2018graph}.  \gat  modifies the aggregation operation in \eqref{eq:conv} by introducing attention weights over the edges. In particular, the $p$-th attention operation results in the following aggregation operation, where 
\begin{equation}
\label{eq:gat}
    \boldsymbol{z}_{i}^{(\ell,p)}\leftarrow \sum_{j \in \mathcal{N}_{i}\cup i} \alpha_{i j}^{p}  \boldsymbol{x}_{j}^{(\ell-1)},
\end{equation}
where $\alpha_{i j}^{p}$ are normalized attention coefficients computed by the p-th attention mechanism. In the transformation step, the $P$ intermediate representations corresponding to $P$ attention mechanisms are concatenated after a non-linear transformation as in \eqref{eq:transgcn} to obtain the representation at layer $\ell$.
\begin{equation}
\label{eq:gat}
    \boldsymbol{x}_{i}^{(\ell)}=||_{p=1}^{P} \operatorname{ReLU}\left( \boldsymbol{z}_{i}^{(\ell,p)} \mathbf{W}^{(p\ell)} \right),
\end{equation}
where $||$ denotes concatenation operator and $\mathbf{W}^{(p\ell)}$ is the corresponding weight matrix at layer $\ell$. For more details of the attention computations, we refer the reader to \cite{velickovic2018graph}.

\mpara{GraphSage (\graphsage)} \cite{hamilton2017inductive}. GraphSage generalizes the graph convolutional framework by proposing several aggregation operations. To achieve scalability, rather than using the complete 1-hop neighborhood of the node, it samples a fixed number of neighbors randomly at each layer for each node. Let $\mathcal{\tilde{N}}(i)$ denote the set of sampled neighbors for node $i$. The  aggregation (we use mean aggregation in this work) operation as applied in \graphsage is given as follows.
\begin{equation}
\label{eq:sage}
\boldsymbol{z}_{i}^{(\ell)} \leftarrow \operatorname{CONCAT}\left(\boldsymbol{x}_{i}^{(\ell-1)},
\frac{1}{|\mathcal{\tilde{N}}(i)|} \sum_{j\in \mathcal{\tilde{N}}(i)} \boldsymbol{x}_{j}^{(\ell-1)}\right).
\end{equation}
The transformation operation stays the same as in \eqref{eq:transgcn}.

Our approach is the first work to compare different graph convolution-based models with respect to their vulnerability to MI attack. More precisely, we ask \emph{if the differences in the aggregation and transformation operations of the graph convolution-based models lead to differences in privacy risks.} 

\subsection{Privacy attacks on Machine Learning}

Several attacks on machine learning models have been proposed including membership inference attack \cite{Shokri17, carlini2019secret, Yeom17, nasr2018machine, salem2018ml} where the adversary aims to infer whether a data sample was part of the data used in training a model or not. In the attribute inference attack, the attacker's goal is to reconstruct the missing attributes given partial information about the data record and access to the machine learning model \cite{zhao2019inferring, Yeom17, jayaraman2019evaluating}. In model inversion attack \cite{fredrikson2015model, melis2019exploiting, papernot2016towards}, the model’s output is used to extract features that characterize one of the model’s classes. The goal of model extraction and stealing attack is to steal model parameters and hyperparameters to duplicate or mimic the functionality of the target model \cite{Tram10.5555/3241094.3241142, wang8418595}. However, little attention has been paid to the privacy risks of GNNs.
Recently, privacy preserving learning algorithms for GNN models have been proposed \cite{sajadmanesh2020differential, zhou2020privacy}. However. their proposed solutions are not directly applicable in overcoming the risk incurred by MI attacks. 
After our work, a recent paper on node-level membership inference attack for GNN was proposed \cite{he2021node}. Their work is different from our work in that we focus on analyzing  the  properties  of  GNNs  and dataset properties that determines  the  differences  in  their  robustness. Moreover, as indicated by the authors, their proposed defenses limits the target model's utility whereas our proposed defenses does not affect model's utility.

\section{Our Approach}
\subsection{Problem Description}
 \subsubsection{Notations} Let $G=(V,E)$ represents the graph dataset with $|V|$ nodes and $|E|$ edges. Let the nodes be labeled. We denote by \emph{target graph}, $\mathcal{G}_t=(V_t, E_t)$,  the induced graph on the set of sensitive or the member nodes $V_{t}$ which is used to train the \emph{target model}, $\mathcal{M}$.

\begin{PS}
Let a GNN model $\mathcal{M}$ be trained using the graph $G_t=(V_t, E_t)$. Given a node $v$ and its $L$-hop neighborhood, determine if $v \in V_t$. Note that even if $v$ was in the training set, the $L$-hop neighborhood known to the adversary might not be the same as the one used to train the model $\mathcal{M}$. 
\end{PS}

\subsubsection{Our Proposed Settings} We  propose two realistic settings for carrying out MI attack on GNNs: (i) in the first setting which we call the \textit{TSTF (train on subgraph, test on full) setting}, in which the whole graph $G$ is available to the adversary but she is not aware of the subgraph $\mathcal{G}_t$ used for training the target model. This implies that the attacker has access to the links (if any) between the member nodes and non-member nodes (ii) in our second setting \textit{TSTS (train on subgraph, test on subgraph) setting}, the target graph $G_t$ is an isolated component of $G$, i.e., the member and non-member nodes are not connected. The adversary has access to $G$ but does not know which of its component is used for training the target model.

\begin{figure}
\centering
\includegraphics[width=\linewidth]{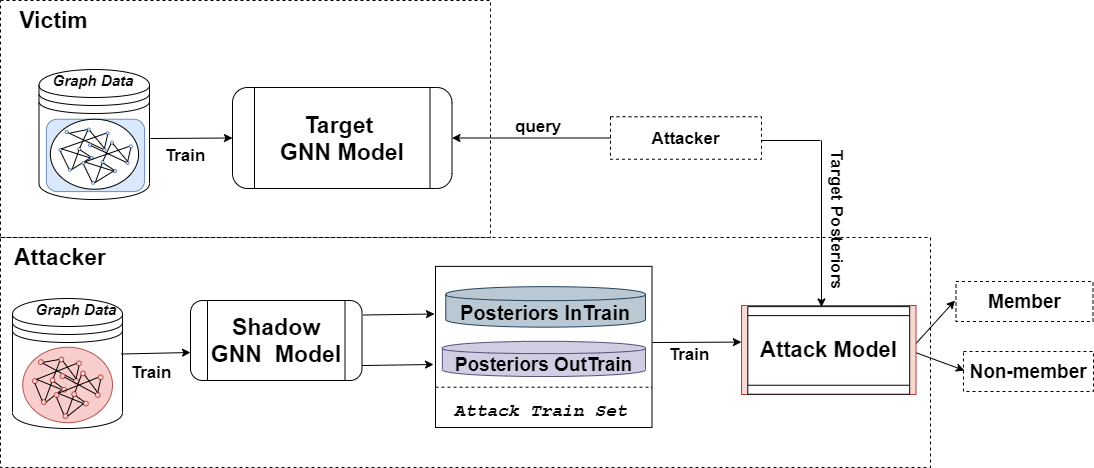}
\caption{\label{fig:mia-attack-graph} Attack methodology for membership inference on GNN models. The training nodes and neighbor information used for training the shadow GNN model are labeled as \textit{Member}. We also query the shadow model with nodes from a test graph and labeled the predictions as \textit{non-Member}. These predictions are used to train the attack model. The attacker then queries his trained attack model with posteriors obtained from the target model (target predictions) to infer membership.}

\end{figure}
\subsection{Attack Methodology} We model the problem of membership inference as a binary classification task where the goal is to determine if a given node $v\in V_t$. We denote our attack model by \(\mathcal{A}\). 

We organize the adversary's methodology (also shown in Figure~\ref{fig:mia-attack-graph}) into three phases, shadow model training, attack model training, and membership inference.

\subsubsection{Shadow model training}
To train the shadow model, we assume that the adversary has access to the graph with vertex set \(V_s\) that comes from the same underlying distribution as \(V_t\) (the assumption which we also relax in Section \ref{sec:datatransferattack}). Then the adversary trains the shadow model using the shadow model's training split, \(V^{Train}_s \subset V_s\). To replicate the behavior of the target model, we use the output class probabilities of the target model (when \(V^{Train}_s\) is used as input) as the ground truth for training the shadow model.
This would result in querying the target model for each vertex in \(V^{Train}_s\). We later relax the number of queries required to $0$ by directly training the shadow model on the original ground truth labels of \(V^{Train}_s\). We observe there is no significant change in attack success rate (c.f. Section~\ref{sec:queries}).
We also find that we do not need to know the exact target model. In fact, we show that using \gcn as the shadow model, irrespective of the actual target model already results in good attack performance (c.f. Section~\ref{sec:targetmodel}).

\subsubsection{Attack model training}
To construct the attack model, we use the trained shadow model to perform predictions over all nodes in \(V^{Train}_s\) and \(V^{Out}_s = V_s \setminus V^{Train}_s\) and obtain the corresponding output class posterior probabilities. 
For each node, we take the posteriors as input feature vectors for the attack model and assigns a label 1 if the node is in \(V^{Train}_s\) and 0 if the node is from \(V^{Out}_s\). These assigned labels serve as ground truth data for the attack model. All the generated feature vectors and labels are used in training the attack model.

\subsubsection{Membership inference}
To perform the inference attack on whether a given node \(v\in V_t\), the adversary queries the target model with \(v\) and its known neighborhood to obtain the posteriors. Note that even if $v$ was part of training data, the adversary would not always have access to the exact neighborhood structure that was used for training.
Then she inputs the posteriors into the attack model \(\mathcal{A}\) to obtain the membership prediction.

\section{Experiments}
\label{sect:experiments}
We compare four popular GNN models: (i) graph convolution network (\gcn)\cite{kipf2017semi}, (ii) graph attention network (\gat) \cite{velickovic2018graph} (iii) simplified graph convolution (\sgc) \cite{pmlr-v97-wu19e} and (iv) GraphSage ( \graphsage) \cite{hamilton2017inductive} as explained in Section \ref{sec:gnns}.
We ran all experiments for 10 random data splits (i.e., the target graphs, shadow graphs as well as test graphs were generated 10 times) and report the average performance along with the standard deviation.

\subsection{Dataset and Settings}
\label{sect:datasetandsettings}
To conduct our experiments, we used 5 different datasets commonly used as a benchmark dataset for evaluating GNN performance. The properties of the datasets are shown in Table \ref{tab:data-stat}.

\begin{table}
\footnotesize 
\caption{\label{tab:data-stat} Dataset Statistics. $|V|$ and $|E|$ denote the number of vertices and edges in the corresponding graph dataset. $|V_t|$ denotes the number of vertices in the target/shadow (train) graph. $deg$ is the average degree of $|V_t|$ calculated for the training graph of target model. The average degree for trainset for shadow model $|V_s|$ is approximately the same as for the target model.}
\begin{tabular}{llllll}
\toprule

& \textbf{Cora} & \textbf{CiteSeer} & \textbf{PubMed} & \textbf{Flickr} & \textbf{Reddit} \\
\midrule

$|V|$    & 2708     & 3312        & 19,717           & 89,250           & 232,965          \\
$|E|$   & 5429      & 4715    & 44,338           & 449,878          & 57,307,946        \\
\textbf{\# features}   & 1433   & 3703    & 500             & 500             & 602             \\
\textbf{\# classes}  & 7                 & 6             & 3               & 7               & 41              \\
\textbf{$|V_{t}|$}     & 630               & 600          & 4500           & 10,500           & 20,500            \\
\textbf{$deg_{TSTS}$}     & 1.111               & 0.41          & 1.102           & 1.638           & 70.383            \\
\textbf{$deg_{TSTF}$}     & 3.898               & 2.736          & 4.496           & 10.081           & 491.987            \\

\bottomrule
\end{tabular}
\end{table}

\subsection{Model Architecture and Training}

We used a 2-layer \gcn, \gat, \sgc, and \graphsage architecture for our target models and shadow models. The attack model is a 3-layer MLP model. All target and shadow models are trained such that they achieve comparable performance as reported by the authors in the literature.
We vary the learning rates between 0.001 and 0.0001 depending on the model and dataset.

\mpara{Evaluation Metrics.} We report AUROC scores, Precision, and Recall for the attack model as done in \cite{Shokri17,salem2018ml}. For the target GNN models, we report train and test accuracy. Due to space constraints we show in the main paper summarized results using mostly the AUROC metric. The detailed results are shown on our GitHub page \footnote{\label{githubpagefootnote}\url{ https://github.com/iyempissy/rebMIGraph}}.

\subsection{Research Questions}
\label{sec:researchquestions}
Here, we summarize the main research questions that we investigate in this work.
\begin{RQ}
\label{rq:compare}
How do different GNN models compare with respect to privacy leakage of training data? What factors lead to differences in vulnerability of GNN models towards MI attack? (c.f. Sections~\ref{sect:GNNComp} and ~\ref{sec:properties})
\end{RQ}
\begin{RQ}
\label{rq:overfit}
How does \textit{overfitting} influence the performance of MI attacks in GNNs? (c.f. Section \ref{sec:overfitting}) 
\end{RQ}

\begin{RQ}
\label{rq:sensitivity}
How does the number of queries for shadow model training, absence of knowledge of similar data distribution, target model architecture and hyperparameter settings affect the attack performance? (c.f. Section \ref{sec:sensitivity}) 
\end{RQ}

\begin{RQ}
\label{rq:defense}
How could we defend against the blackbox MI attack without compromising the model performance? (c.f. Section \ref{sec:defensemechanisms}) 
\end{RQ}

\section {Analysing the MI Attack on GNNs}
\subsection{Overall Attack Performance}
\label{sect:GNNComp}
In this section, we answer the first part of RQ \ref{rq:compare}.

\begin{figure}[h!]
\centering
\begin{subfigure}{0.49\linewidth}
\includegraphics[width=\linewidth]{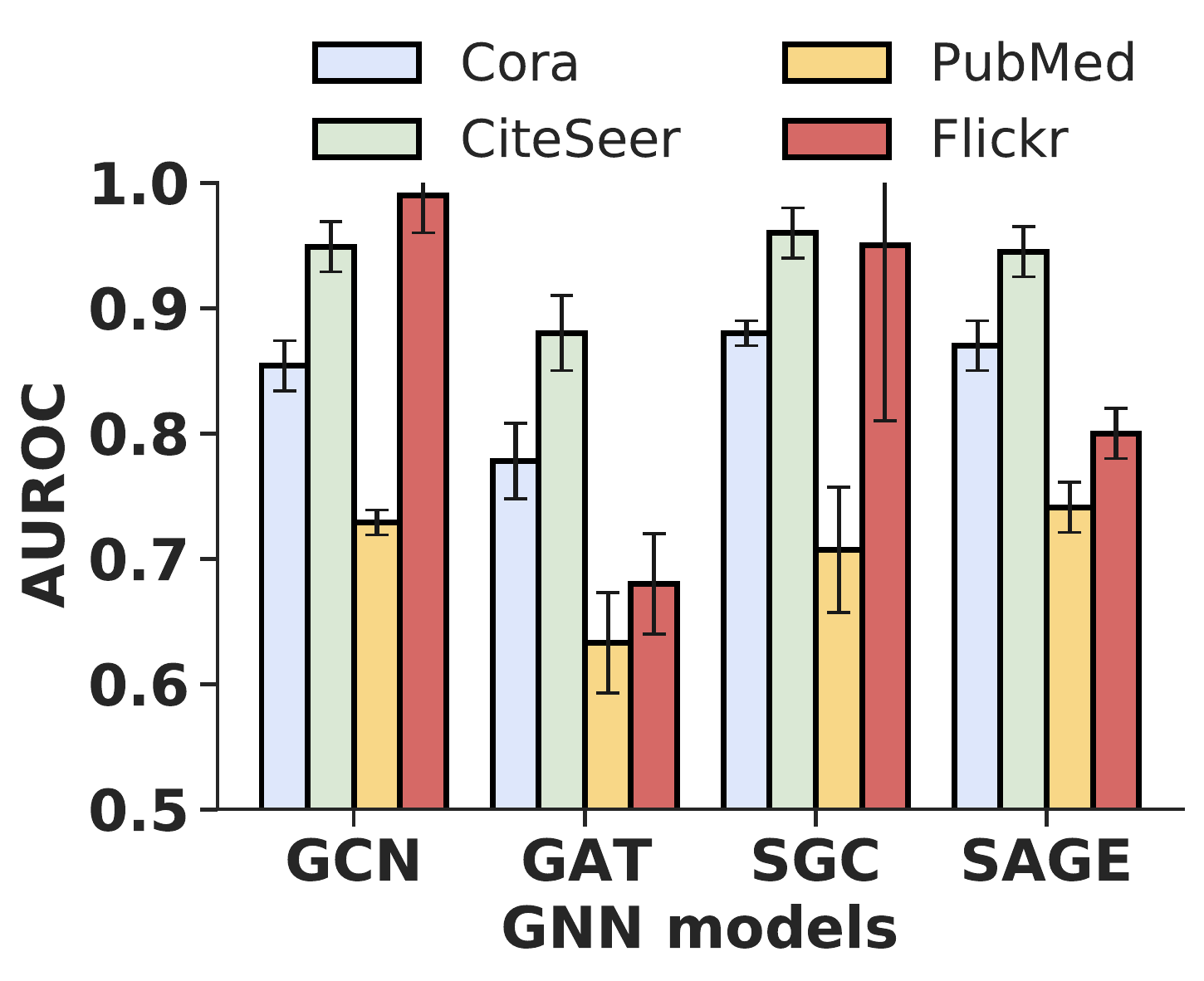}
\caption{TSTF Setting}
\label{AUROC-TSTFResult}
\end{subfigure}
\begin{subfigure}{0.49\linewidth}
\includegraphics[width=\linewidth]{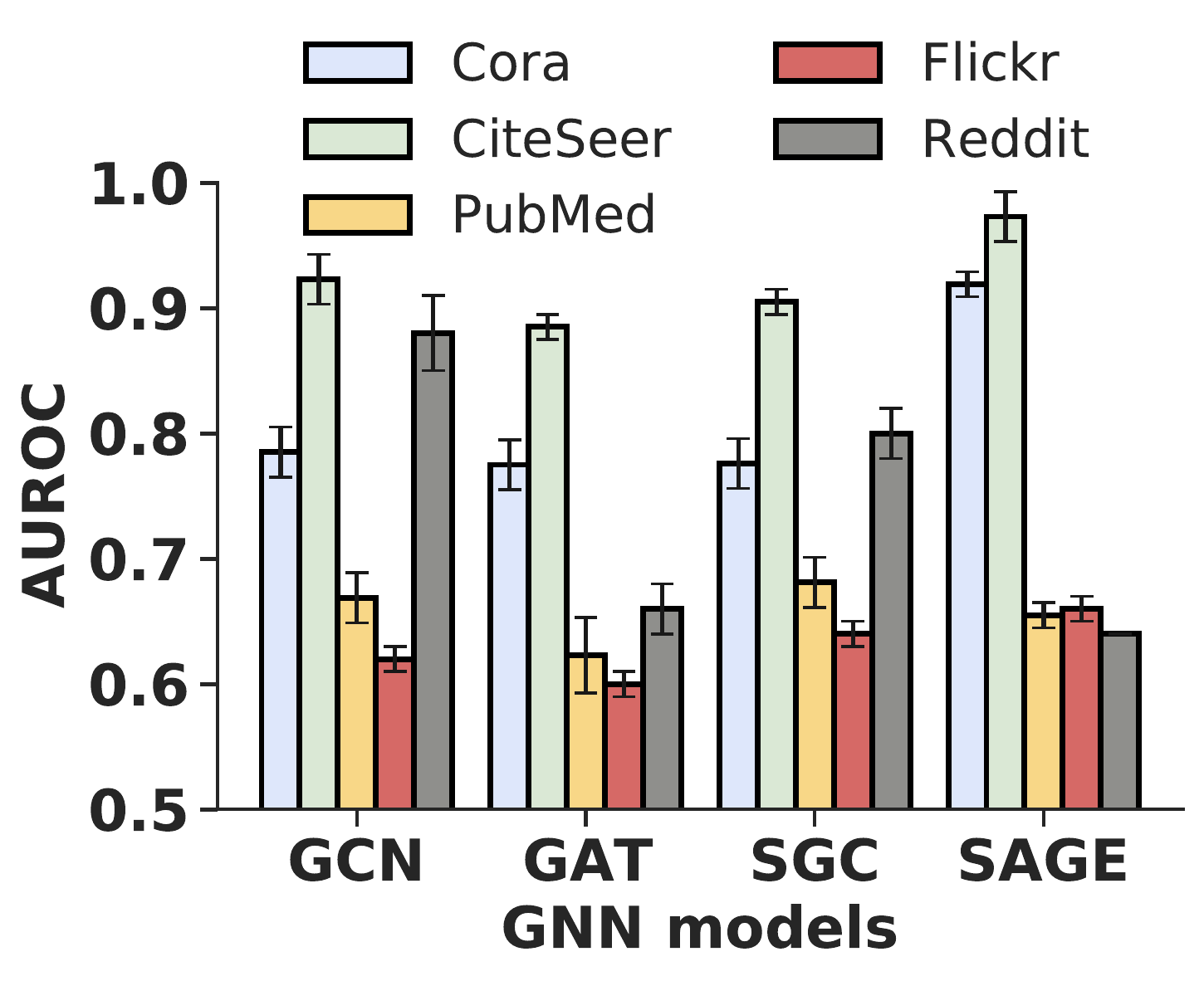}
\caption{TSTS Setting}
\label{AUROC-TSTSResult}
\end{subfigure}

\caption{Mean AUROC scores of attack model against different GNN models. All target models except \graphsage encountered memory issues for the \reddit in TSTF setting. Therefore, we only provide results for \reddit in the TSTS setting.}
\label{fig:compareGNN}
\end{figure}

\paragraph{TSTF Setting} The AUROC scores for the attack model on all datasets except Reddit are shown in Figure \ref{AUROC-TSTFResult}. For the models \gcn and \sgc, the attack model obtains similar scores. Note that the difference between \sgc and \gcn is that \sgc does not use a non-linear transformation after the feature aggregation step. The feature aggregation scheme employed in both models is exactly the same. 

\gat is the most robust towards the attack. \gat also differs from the models in that it uses a weighted aggregation mechanism. 
 \graphsage employs a mean aggregation over the neighborhood's features. Unlike the other models, for the aggregation step, it samples a fixed number of neighbors rather than using the complete neighborhood. Though it shows similar results for 3 citation networks, the attack is less successful for the larger graph \flickr (when compared to \gcn and \gat). We attribute the reason for such an observation to the induced noise in the neighborhood structures because of the random sampling of neighbors. Obviously, the effect is more prominent in  denser graphs like \flickr as compared to \cora where the average degree is less than 2.

\paragraph{TSTS Setting} Unlike in the TSTF setting, the train and test sets in this setting are disconnected. This implies that any node $v\in V_t$ and its exact neighborhood used during training is known to the adversary. We also see a huge reduction in test set performance
implying that the model is not generalizing well to the test set. Intuitively, it would be much easier to attack in this setting.

The AUROC scores for the corresponding attack are shown in Figure~\ref{AUROC-TSTSResult}. Precision and recall of the attack model along with the train-test set performance of the target model are provided in Table VI on Github (\ref{githubpagefootnote}).
We observe that for \cora and \cseer the attack has a similar success rate as in TSTF setting, for \flickr on the other hand, the attack performance degrades. For the larger dataset \reddit, the attack is successful for \gcn and \sgc models with a mean precision of 0.81 and 0.74 
respectively. \gat and \graphsage shows more robustness with AUROC scores close to 0.5 (implying that the attack model cannot distinguish between member and non-member nodes better than a random guess) for datasets: \pubmed, \flickr, \reddit.

\subsection{Effect of Model and Dataset Properties}
\label{sec:properties}
To answer the second part of RQ \ref{rq:compare}, we analyse the differences in the aggregation operation of models and three dataset properties to explain the differences in attack performance.

\subsubsection{Which model is more robust towards MI attack and Why?}
\label{sec:modelprop}
To summarize the above results, we found \gat to be most robust towards membership inference attacks. The reason can be attributed to the learnable attention weights for different edges. The above fact implies that instead of the original graph  model, a distorted one dictated by supervised signals of class labels is embedded in the model. 
This is in contrast with \sgc and \gcn where the actual graph is embedded with equal edge weights. Also, in \graphsage, which uses neighborhood sampling before the aggregation operation, does not use the complete information of the graph during training. The effect is more prominent in denser graphs in which only a small fraction of the neighborhood is used during a training epoch. 

Another interesting observation is the attack behavior changes with datasets. While \gat is overall less vulnerable than other models, the percentage drop in attack performance (as compared to, for example, \gcn) for \flickr (32\%) is much larger than for \cora (9\%). 

\subsubsection{How do dataset properties affect attack performance?}
\label{sec:dataprop}
To investigate the differences in the behavior of the attack model on different datasets we consider three properties of the datasets (i) \emph{average degree} which influences the graph structure and the effect of aggregation function of the GNN (ii) the \emph{number of input features} that influence the number of model parameters and (iii) the \emph{number of classes} that decides the input dimension/features for the attack model.

First, note that for very low average degree graphs the effect of aggregation operation is highly decreased as there would be very few or no neighbors to aggregate over. 
From Table \ref{tab:data-stat}, we observe that \cseer has the lowest average degree (both in the TSTF and TSTS settings) leading to similar attack vulnerability of all GNN models. \reddit, on the other hand, with the highest average degree exhibits a high vulnerability to the attack when \gcn and \sgc are the target models whereas fpr \gat and \graphsage attack performance drastically reduces owing to reasons discussed in the last section. Similar observations can be made for \flickr which has the second-highest average degree. Differences in attack performance for \flickr are smaller as compared to \reddit. This is expected as \reddit has an average degree which is around 50 times the average degree of \flickr for TSTF setting and around 70 times for TSTS setting.

Second, for the three datasets \cora, \cseer, and \pubmed, which exhibit similar average degrees, attack performance is highest for \cseer followed by \cora. The trend stays the same for different target models.
The same pattern is also observed in the number of input features. While \cseer has the highest number of features, \pubmed has the least. Note that the number of input features leads to an increase in the number of parameters (the number of parameters corresponding to the first hidden layer will be $p\times h$ where $p$ is the number of input features and $h$ the hidden layer dimension). The higher number of parameters, in turn, leads to better memorization by models, which explains the above-observed trend in low average degree datasets. 

Third, we recall that the output posterior vector is the input feature vector for the attack model. As the dimension of the posterior vector is equal to the number of classes, more information is revealed for datasets with larger number of classes. The low attack performance on \pubmed can be therefore additionally attributed to its lowest number of classes.

\subsubsection{Effect of Neighborhood Sampling in \graphsage}
\label{sec:neigborsampling}
We attribute the differences in \graphsage's robustness towards attacks on different datasets to its neighborhood sampling strategy. Recall that rather than using complete neighborhood in the aggregation step, \graphsage samples a fixed number of neighbors at each layer. SAGE also utilizes a mini-batching technique that contains nodes on which representation needs to be generated and their sampled neighbors.
To showcase the effect of the neighborhood sampling, we varied the number of neighbors sampled at different layers of the network and the batch size. 

\begin{figure}[h!]
\centering
\begin{subfigure}{0.49\linewidth}
\includegraphics[width=\linewidth]{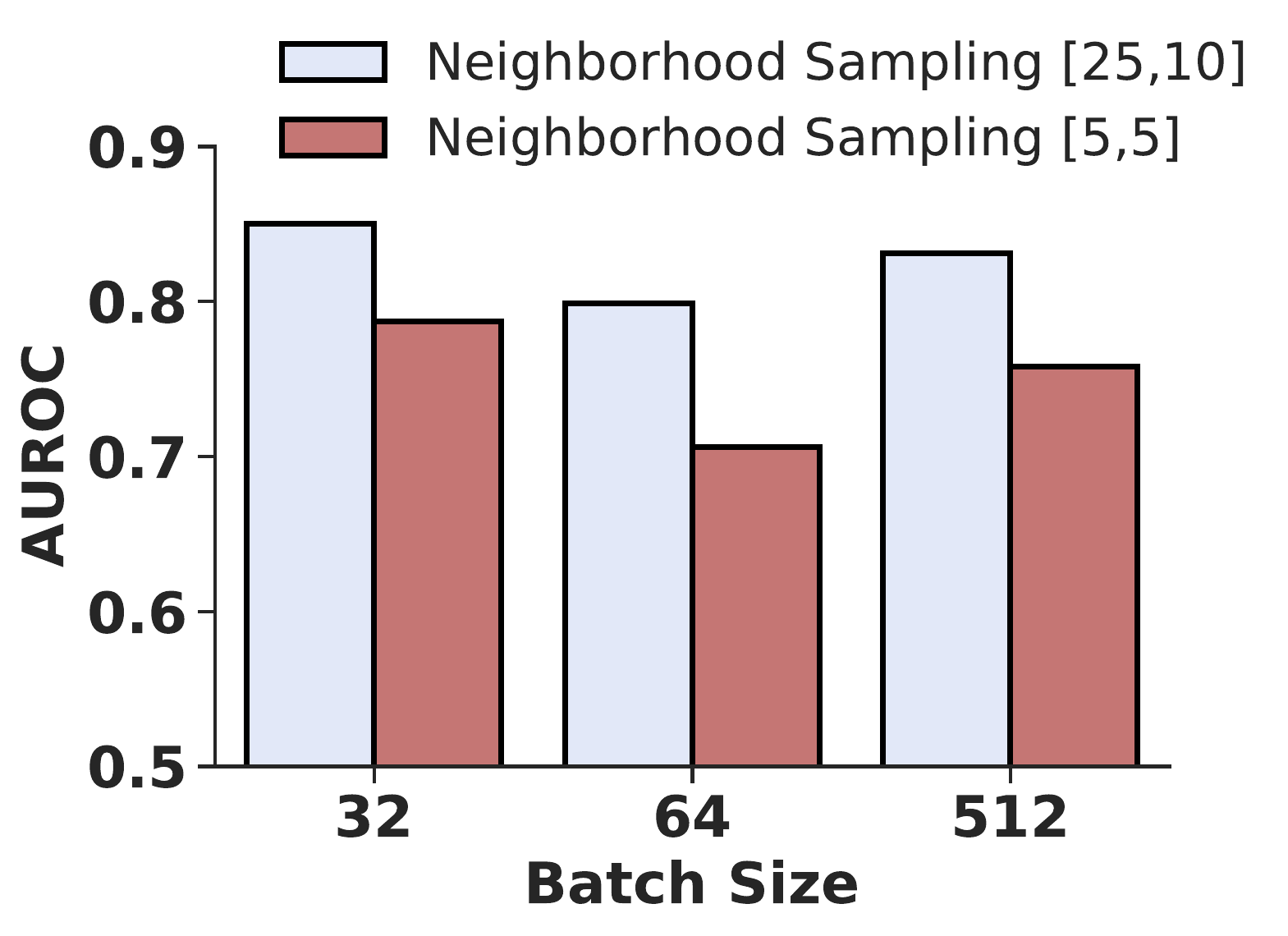}
\caption{\cora}
\label{fig:sampleCora}
\end{subfigure}
\begin{subfigure}{0.49\linewidth}
\includegraphics[width=\linewidth]{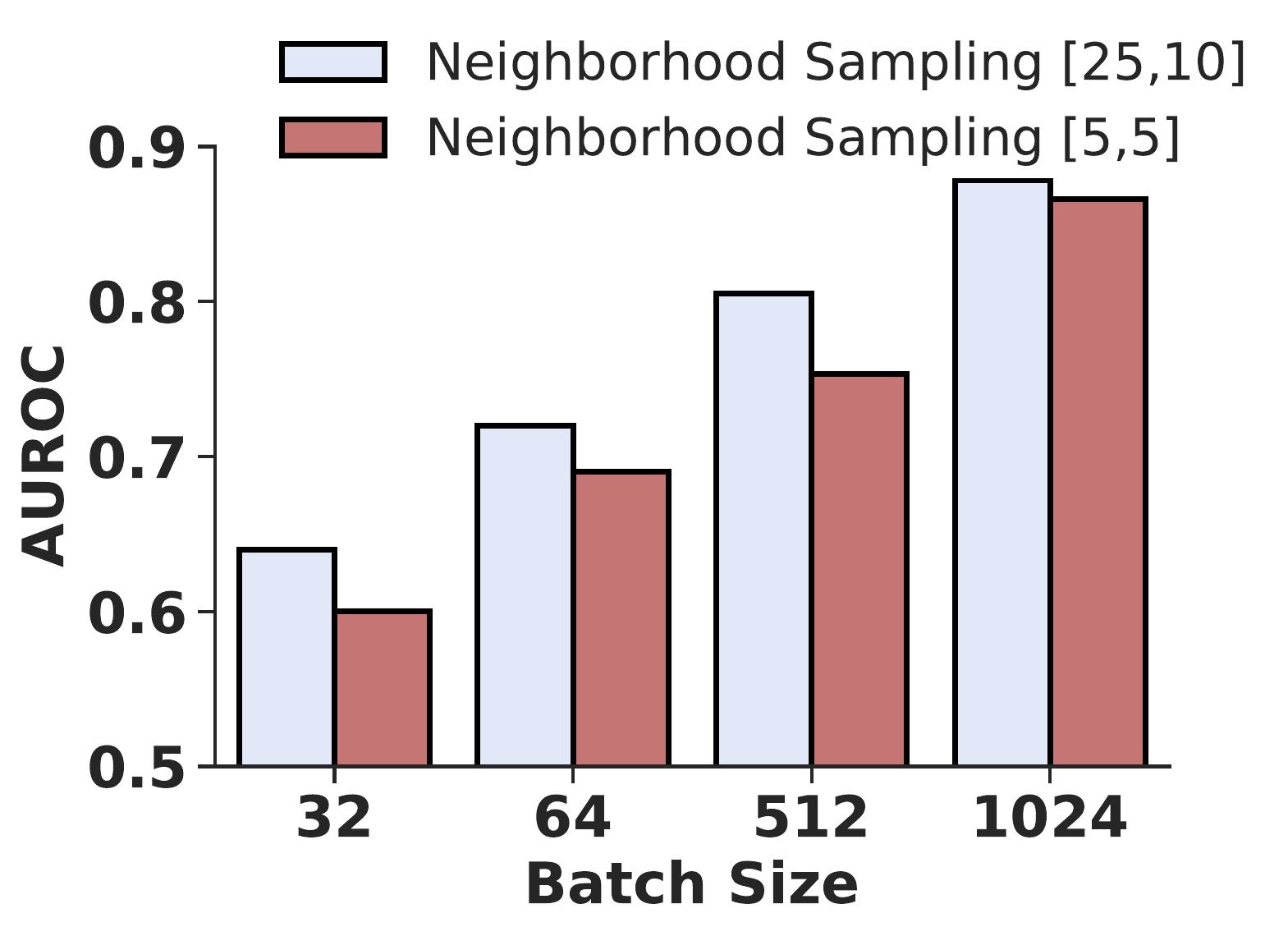}
\caption{\flickr}
\label{fig:sampleFlickr}
\end{subfigure}

\caption{\label{BatchSizevsNeighborSage} Effect of training batch size and sampled neighbors on attack performance for \graphsage model on (a) \cora and (b) \flickr dataset.  }
\end{figure}

We used [25,10] and [5,5] as sampled neighborhood sizes in layers 1 and 2. 
As shown in Figure \ref{fig:sampleCora}, the attack AUROC decreases as the number of sampled nodes decreases. This is because the model uses the noisy neighborhood information and it is not able to fully encode the graph structure in the model, this, in turn, makes the posteriors of neighboring nodes less correlated. Similar results are obtained for a larger dataset, \flickr (shown in Figure~\ref{fig:sampleFlickr}).

\subsubsection{Effect of Instance Connectivity}
\label{sec:instanceconn}
Here, we present a qualitative analysis of the differences in the robustness of different models to MI attack using \flickr as an example dataset in the TSTF setting. 
 Recall that given the predicted posteriors as input, the attack model labels the node instance as a member (label 1) or non-member (label 0) node. To understand the pattern of label assignments by the attack model we need the following definition.
\begin{definition}[Homophily] \label{def:homophily} For any node $u$ which is either a member or non-member, we define its homophily as the fraction of its one-hop neighbors which has the same membership label as $u$. The neighborhood of any node is computed using the graph available to the adversary. We call homophily with respect to ground truth as the \textbf{true homophily} and with respect to the attack model predictions as the \textbf{predicted homophily}.
\end{definition}

\begin{figure*}[t!]
\centering
\includegraphics[width=0.85\linewidth]{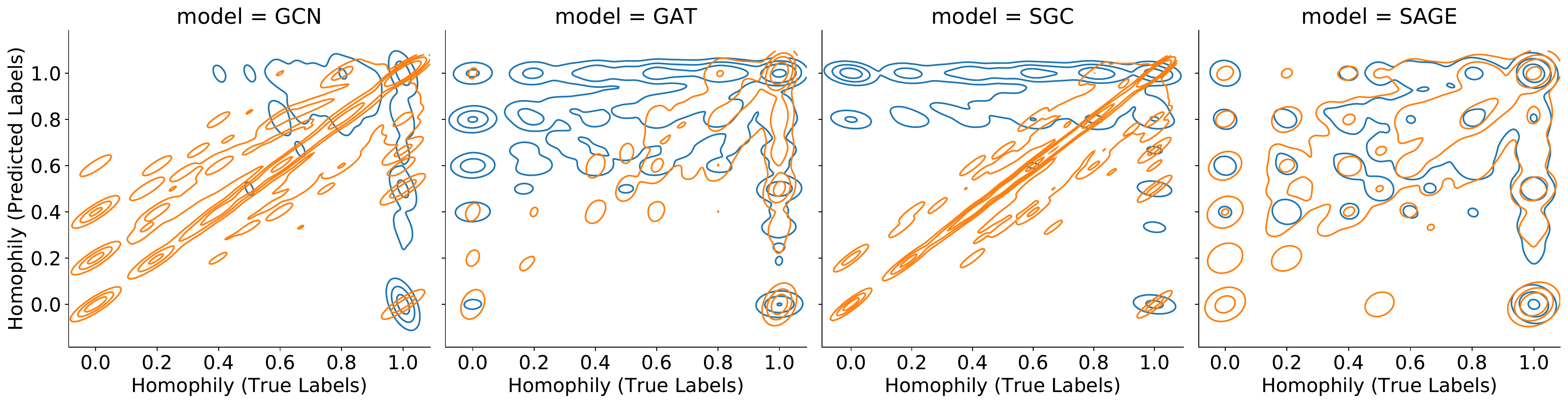}

\caption{Joint density plot of true and predicted homophily (refer to Definition \ref{def:homophily}) on \flickr dataset.}
\label{fig:flickrHomophily}
\end{figure*}

Therefore, true homophily of $1$ means $u$, and all its neighbors in the graph used by the adversary have the same membership label. Similarly, predicted homophily of $1$ implies that $u$ and its neighbors were assigned the same membership label by the attack model. In Figure \ref{fig:flickrHomophily}, we visualize the differences in attack behavior for different models on the \flickr dataset by plotting the joint distribution of true and predicted homophily of the correctly (orange contour lines) and incorrectly (blue contour lines) predicted nodes. We chose \flickr here because the attack performance varies the most with respect to different target models as also discussed in the last sections. 

We observe more dense regions in the upper half of the plots for all the models. Noting the fact that these highly concentrated regions correspond to high predicted homophily, we conclude that the attack model's predictions on a node are highly correlated with its predictions on its neighbors. As the attack model is agnostic to the graph structure, this further implies that the posterior of neighboring nodes are also correlated, which the attack model can exploit. 

The differences in the behavior of different models are also well illustrated. Note that the higher the density of orange regions on diagonals (see for \gcn and \sgc), the more accurate the attack model will be. In contrast to \gcn,  the attack model is confused for \gat and assigned the wrong label to corresponding nodes and their neighbors (see blue regions corresponding to high predicted homophily). For \graphsage, even though there are more orange regions, these do not lie over the diagonal. This means that the attack model, even if it predicts the right membership label for a member node, it also predicts the same membership label for its non-member neighbors. Hence, including them incorrectly in the member set.  To summarize, for \gat the attack results in more false negatives whereas for \graphsage there are more false positives. Both scenarios render the attack less useful to the adversary.

\subsection{Effect of Model Overfitting}
\label{sec:overfitting}
To investigate the effect of overfitting (RQ \ref{rq:overfit}), we train the models such that they achieve zero training loss or high generalization error. The train and test accuracy as well as the attack precision and recall are  shown in Table \ref{overfit-table}. 

\begin{table}[h!]
\footnotesize
\caption{\label{overfit-table} Performance of GNN models and the attack model on \flickr dataset in case of overfitting.}
\begin{tabular}{lccccc}
  
 \toprule
   & \multicolumn{2}{c}{\textbf{Target Mode}l} && \multicolumn{2}{c}{\textbf{Attack Model}} \\
 & \textsc{Train} & \textsc{Test} && \textsc{Precision} & \textsc{Recall}\\
 \midrule
\gcn &  $0.70 \pm 0.01$ & $0.13 \pm 0.01$ && $0.63 \pm 0.01$ & $0.70 \pm 0.04$  \\
\gat &  $0.61 \pm 0.07$ & $0.50 \pm 0.21$ && $0.75 \pm 0.05$ & $0.69 \pm 0.03$  \\
\sgc &  $0.75 \pm 0.02$ & $0.20 \pm 0.03$ && $0.62 \pm 0.04$ & $0.59 \pm 0.09$  \\
 \graphsage &  $0.90 \pm 0.03$ & $0.24 \pm 0.03$ && $0.57 \pm 0.08$ & $0.50 \pm 0.01$  \\
\bottomrule
\end{tabular}
\end{table}

\begin{figure}[h!]
\centering
\begin{subfigure}{0.49\linewidth}
\includegraphics[width=\linewidth, height=4.9cm, keepaspectratio]{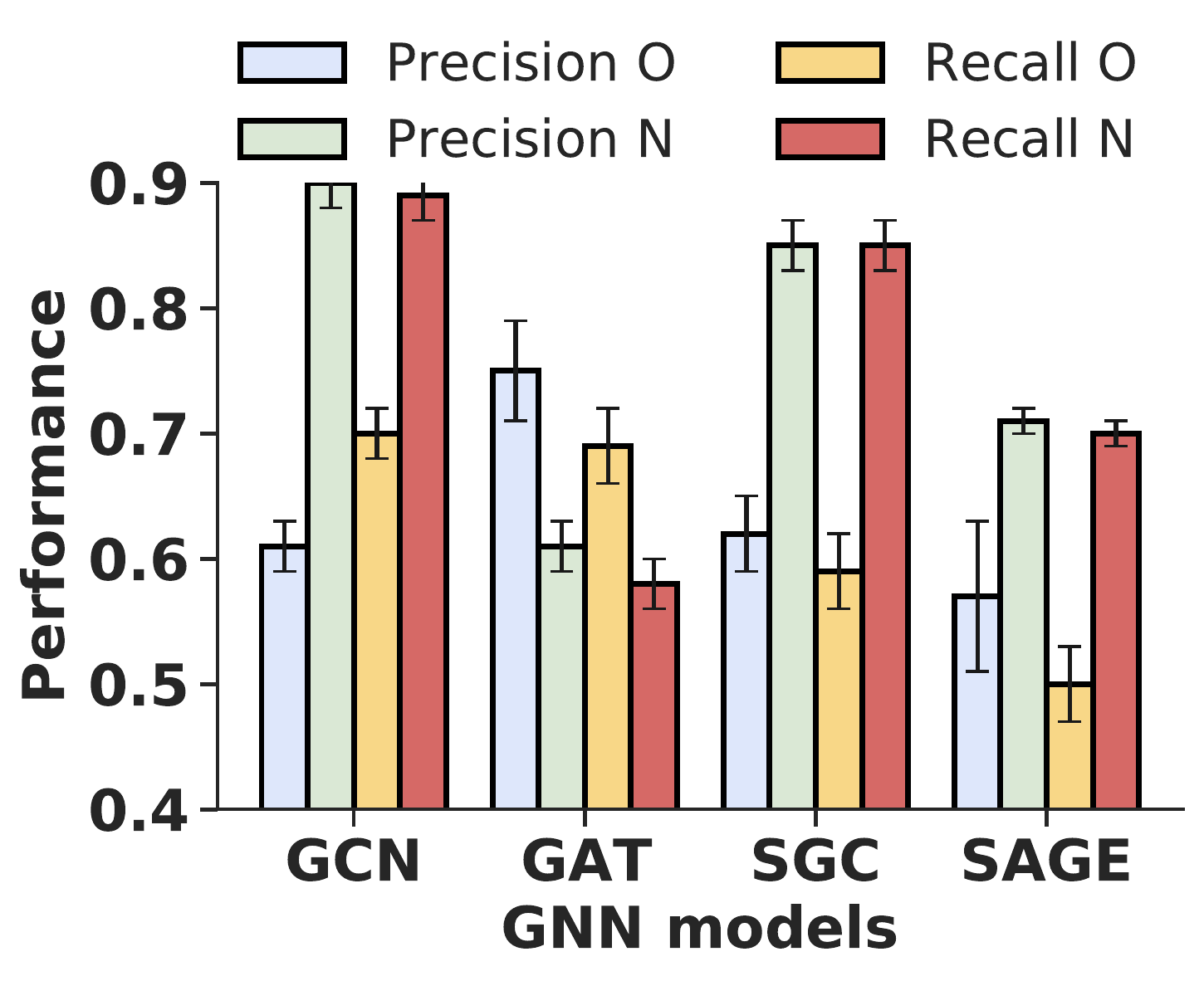} 
\caption{ \footnotesize{Influence of overfitting on the attack on \flickr. N= Normal and O=Overfit.  "\textbf{Normal}" refers to the case when all models were trained for a fixed number of epochs.} }
\label{Overfitting-effect}
\end{subfigure}
\hfill
\begin{subfigure}{0.49\linewidth}
\centering
\includegraphics[width=\linewidth]{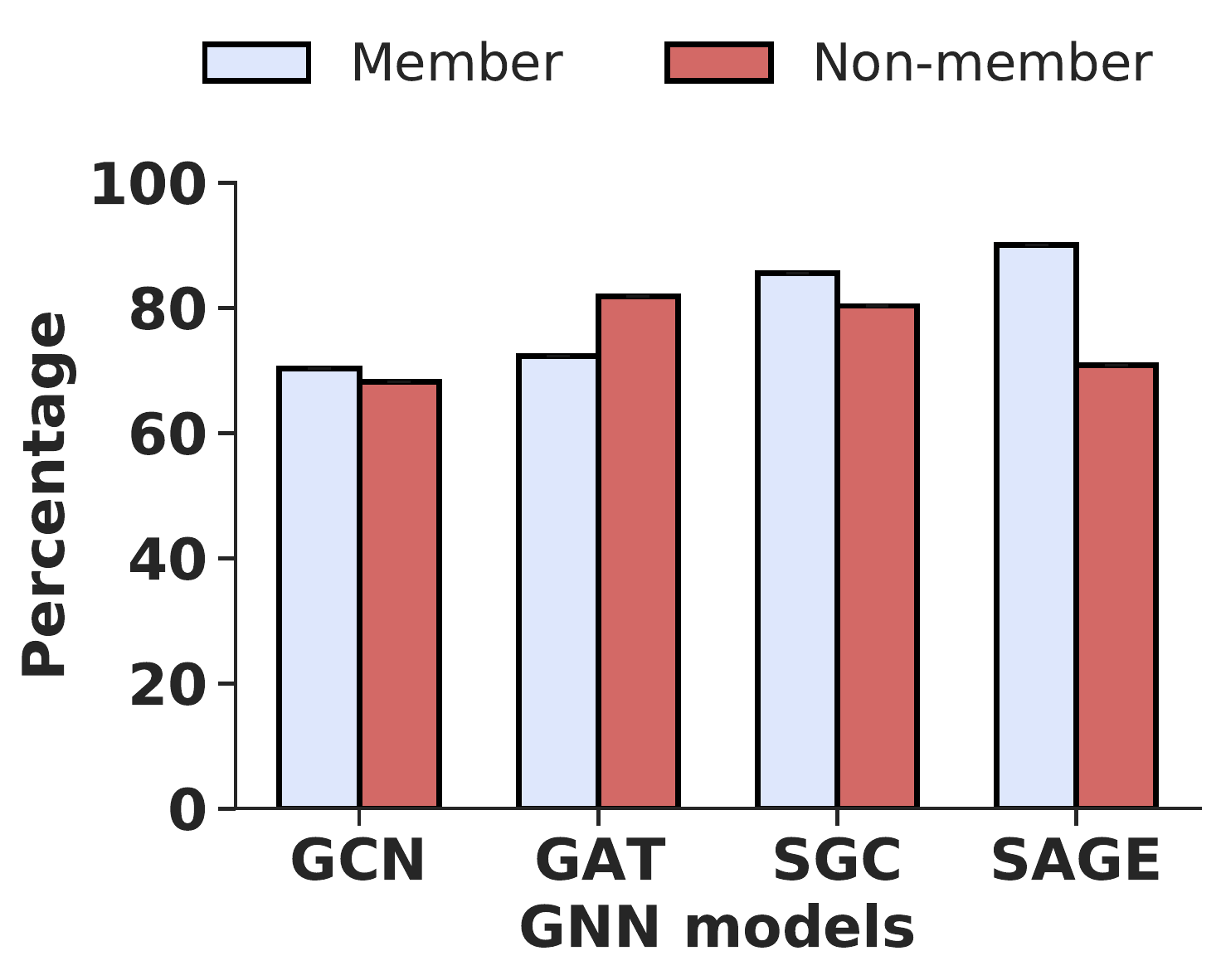} 
\caption{ \footnotesize{\% of nodes for which the maximum posterior is greater than 0.8 for overfitted GNN models. The statistics are shown for one random data split.}}
\label{overfittingDistribution}
\end{subfigure}
\caption{Effect of overfitting. In figure (a) we observe a surprising effect that attack is less successful for overfitted models. The reason behind such an effect is explained in (b) which illustrates that the overfitted model also makes extremely confident albeit incorrect predictions on the test (unseen) set.}
\end{figure}

Figure \ref{Overfitting-effect} shows the comparison between a "normal" model and the overfitted model. The attack precision and recall of the overfitted model consistently decreases across all models except for GAT. 
This implies that overfitting alone might not always be a contributing factor to membership inference attack and that overfitted model may not always encode the information needed to launch an MI attack. 

To understand the reasons behind the above observations, we investigate the posterior distribution of member and non-member nodes. 
In Figure \ref{overfittingDistribution}, we show the distribution of the \textit{maximum} posterior (i.e., the posterior probability corresponding to the predicted class) of overfitted models on the members and non-members. We observe that in the case of overfitting, the GNN model not only makes highly confident predictions for the member nodes but also for the non-members. Most of the nodes whether member or non-member obtain the maximum class probability (or posterior) greater than $0.8$ for models \gcn and \sgc. For \gat and \graphsage, the attack model obtains higher precision given that a relatively less number of non-member nodes obtain a high maximum posterior. Moreover, from the test set performance in Table~\ref{overfit-table}, we observe that \graphsage generalizes better than \gat which also reflects in lower attack precision in \graphsage.

\subsection{Sensitivity Analysis of Attack}
\label{sec:sensitivity}
We answer  RQ\ref{rq:sensitivity} by performing the sensitivity analysis of the attack with respect to the number of queries (Section~\ref{sec:queries}), different sizes of hidden layers (Section~\ref{sec:hyperparameters}) and relaxation of model architecture and data distribution assumptions (Section \ref{sec:targetmodel} and \ref{sec:datatransferattack}).

\subsubsection{Attack with reduced number of queries}
\label{sec:queries}
We relax the number of queries required to imitate a target model to $0$ by assuming that the adversary has access to the dataset from a similar distribution as the dataset used for the target model. To construct such  datasets we randomly sampled disjoint sets of nodes from the full graph for the target as well as the  shadow model. We then construct the corresponding induced graphs on the node sets to train the shadow and target models. Note that the shadow model data, in this case, will not be exactly from the same distribution as the target graph since our construction would not exactly preserve the structural characteristics of these graphs e.g. degree distribution. The data used in training the shadow model is in fact, similar but not from the same distribution as the target model. 
We found that training the shadow model using ground truth labels performs similarly to querying the target model in the order of $\pm 0.02$ standard deviation.

\subsubsection{Attack performance without knowledge of exact  hyperparameters}
\label{sec:hyperparameters}

In this section, we relax the assumption that the attacker knows the exact hyperparameters used in the target model by varying the number of hidden neurons of the shadow model. We experiment with three values $\{256, 128, 64\}$.

The corresponding mean AUROC scores are plotted in Figure 7 on Github  (\ref{githubpagefootnote}) due to space constraint.
A general trend is that the larger the hidden layer size, the better the attack performance. This is expected as an increase in the size of the hidden layer increases the model parameters/capacity to store more specific details about the training set. 
Therefore, though we observe some reduction in attack performance for \pubmed when using 128 or 64 as hidden layer size, an attacker can just choose the hyperparameter which gives the best train set performance on its shadow dataset.

\subsubsection{Attack without the knowledge of target model's  architecture }
\label{sec:targetmodel}
We further relax the assumption that the attacker knows the architecture of the target model. Specifically, we used \sgc as the shadow model and other GNN models as the target model. As the \sgc model is obtained from the \gcn model by removing the non-linear activation function from \gcn, we aim to quantify how this difference affects the attack performance. Therefore, we also used \gcn as a shadow model. The mean AUROC scores corresponding to attacks for different datasets are presented in Figure 8 in our Github page (\ref{githubpagefootnote}).

In both TSTF and TSTS, on the \cseer and \cora dataset, the performance of using different shadow models is equivalent to using the same model as the target model except for \graphsage where a significant drop in performance is observed. 
However, \gcn performs significantly better than \sgc when used as the shadow model by the attacker.
On the \pubmed dataset, an interesting observation, particularly for \gat is that when \sgc is used for the shadow model, the attack precision and recall increases more than when \gat (target model) is used as the shadow model.
On the \flickr and \reddit datasets, using \gcn as the shadow model performs comparably to an adversary knowing the architecture of the target model in both TSTS and TSTF settings. However, using \sgc as a shadow model significantly led to reduced attack precision in the TSTF setting. Better attack precision is achieved when \gcn is used as the shadow model and \graphsage is used as the target model on large networks like \reddit. Therefore, we conclude that using GCN as the shadow model is sufficient to launch a successful attack and that the removed non-linear activation function of \sgc makes it a less attractive option to use as a "universal" shadow model.

\subsubsection{Attack using different data distribution (Data transferring attack)}
\label{sec:datatransferattack}
We relax the assumption that the attacker trains her shadow model based on data coming from similar distribution as that used by the target model. Specifically, we used \cora as the data for training the target model and \cseer as the data used by the attacker for training her shadow model. In here, the goal of the attack model is to understand membership status based on the posterior distribution. To cater for the discrepancies in the length of the posterior vectors of these two datasets, we select the top $n$ coordinates of the posterior vector and arrange them in ascending order. 

As shown in Table \ref{table:relaxdataset}, we observe that relaxing the knowledge of the dataset distribution does not affect the attack precision. Surprisingly, some gains are observed on GCN and GAT. However, the recall drops by $13\%$ on GCN, $16\%$ on GAT and SGC while the recall on SAGE remains the same. This implies that the assumption of the attacker drawing the shadow dataset from the same distribution as the target model can be relaxed with minimal loss in attack performance.

\begin{table}[!h]
\footnotesize
\caption{Attack precision and recall when Cora is used as the target and CiteSeer as the shadow dataset. The \% change with respect to original performance is shown in the brackets.}
\begin{tabular}{p{1.1cm}|l|l|l|l}
 & \gcn & \gat & \sgc &\graphsage \\ \toprule
\textsc{Precision} & 0.766 (+1\%) & 0.723 (+8\%) & 0.780 (-2\%) &0.798 (-2\%)\\
\textsc{Recall}  &0.653 (-13\%) & 0.552 (-16\%) & 0.663 (-16\%) & 0.792 (0\%) 
\end{tabular}
\label{table:relaxdataset}
\end{table}

\section{Defense mechanisms}
\label{sec:defensemechanisms}
To defend against the current black box attack based on posteriors (RQ \ref{rq:defense}), we note that the defense mechanism should possess the following properties. \emph{First,} given access to only posteriors, the defense should lend indistinguishability among member and non-member nodes without compromising task performance and target model's utility. \emph{Second,} the defense should be oblivious to the attacker. The second property is important for output perturbation-based defense mechanisms such that the added noise cannot be inferred from the released information.
 
Based on the insights gained from our experimental analysis of attack performance for different GNN models and datasets we propose two defense mechanisms : (i)  \emph{query neighborhood sampling defense} (\nsd) and (ii) \emph{Laplacian binned posterior perturbation} (\lbp) which we describe in the following sections.
\paragraph{Laplacian binned posterior perturbation (\lbp) defense}
Here, we propose an output perturbation method by adding noise to the posterior before it is released to the user. A simple strategy would be to add Laplacian noise of an appropriate scale directly to each element of the posterior. We refer to this strategy as \vanpd. Note that the noise level increases with the number of classes which can have an adverse effect on model performance. 

To reduce the amount of noise needed to distort the posteriors, we propose a binned posterior perturbation defense.
We first randomly shuffle the posteriors and then assign each posterior to a partition/bin. The total number of bins, $\psi$, is predefined and depends on the number of classes. For each bin, we sample noise at scale $\beta$ from the Laplace distribution (\lbp). The sampled noise is added to each element of the bin. After the completion of the noise addition operation to each bin, we restore the initial positions of the noisy posterior $\textbf{y*}$ before binning. Then we release $\textbf{y*}$.

We observe in our experiments that it leads to a drop in attack performance without substantially compromising model performance on the node classification task. We set the reference values for $\beta$ as \{5, 2, 0.8, 0.5, 0.3, 0.1\}. The higher the value of $\beta$, the higher the added noise. We set $\psi$ as \{2, 3, 4\} where for example, $\psi=2$ implies that the posterior vector is divided into $2$ groups and the same noise added to all members of the same group.

\paragraph{Query neighborhood sampling defense (\nsd)}
Exploiting  the observation that a node and its neighbors are classified alike by the attack model (homophily property), we propose a query neighborhood sampling (\nsd) defense mechanism to distort the similarity pattern between the posterior of the node and that of its neighbors. Specifically, when a target model is queried with the node and its $L$-hop neighborhood, the defender removes all its first-hop neighbors except $k$ randomly chosen neighbors (Note that no change is made to the trained model.). The neighborhood of the $k$ sampled neighbors stays intact and is not changed. 
By doing so, \nsd limits the amount of information used to query the target model. We set the reference values for $k$ as follows \{0, 1, 2, 3\} which implies sampling no neighbors, $1, 2, \text{ or } 3$ neighbors respectively. 

\subsection{Evaluating Defenses}
We measure the effectiveness of a defense by the drop in attack performance after the defense mechanism is applied. To further incorporate the two desired properties of the defense into our evaluation we employ the following utility measures  \cite{jia2019memguard}.

\paragraph{ Label loss ($\mathcal{L}$)} The label loss measures the fraction of nodes in the evaluation dataset whose label predictions are
altered by the defense. 
For a given query $i$, if the highest coordinate of the true posterior and that of the perturbed or distorted posterior is the same, then the $\mathcal{L}_i$ is $0$, otherwise, it is $1$.
The total label loss is quantified as:
$
    \mathcal{L} = \nicefrac{\sum\limits_{i=1}^{|Q|} \mathcal{L}_i }{|Q|},
$
where $|Q|$ is the number of user queries. A $\mathcal{L}$ close to $0$ is desirable whereas $\mathcal{L}$ close to $1$ indicates that the defense mechanism is relatively bad since it alters the label prediction which directly affects the test accuracy of the target model.  

\paragraph{ Confidence score distortion ($\mathcal{C}$)}
For a given query $i$, we measure the confidence score distortion, $\mathcal{C}_i$  by the distance between the true posterior and the distorted posterior due to the defense mechanism. 
We use Jensen Shannon Distance (JSD) as the distance metric. JSD extends Kullback–Leibler divergence (relative entropy) to compute symmetrical score or similarity between two probability distributions. The total confidence score distortion is given as:
$\mathcal{C} = \nicefrac{\sum\limits_{i=1}^{|Q|} \mathcal{C}_i}{|Q|}$
where $\mathcal{C}_i = JSD(\textbf{y}, \textbf{y*})$ for a given query $i \in Q$.
Ideally, $0$ indicates that both the perturbed posteriors and the true posteriors are the same and $1$ indicates that they are highly dissimilar.

\mpara{Setup.}\label{res:setup} We compare the drop in attack precision with respect to label loss ($\mathcal{L}$) and confidence distortion ($\mathcal{C}$). For each defense and a given value of each parameter ($\beta, \psi \text{ or } k$), we plot the pairs $\{\mathbb{P}, \mathcal{L}\} \text{ and } \{\mathbb{P}, \mathcal{C}\}$ where $\mathbb{P}$ is the attack precision. 
We use the attacker's AUROC as a representative metric but observe a similar trend on the other attack inference performance metrics (precision and recall). We performed our experiments on \gcn since it is the most vulnerable GNN model across all datasets as observed in Section \ref{sect:GNNComp} and \ref{sec:targetmodel}.

\subsection{Results}
\begin{table}[h!]
\caption{Attack precision, recall and AUROC (lower the better) corresponding to different label loss $\mathcal{L}$. \textbf{--} indicates that the defense mechanism does not incur the corresponding label loss.
}
\label{tab:labellossattackperf}

\centering
\scriptsize
\setlength{\tabcolsep}{3.5pt}
\settowidth\rotheadsize{Citese}
\begin{tabular}{cllllllllll} \toprule
& &\multicolumn{3}{c}{\textbf{\vanpd}} & \multicolumn{3}{c}{\textbf{\lbp}} & \multicolumn{3}{c}{\textbf{\nsd}} \\
& {$\mathcal{L}$} & {Prec} & {Rec} & {AUC}   & {Prec}    & {Rec}  & {AUC} & {Prec} & {Rec} & {AUC}       \\ \midrule
\multirow{3}{*}{ \rotcell{\scriptsize \cora}}

& \textbf{0}   & 0.814 & 0.811 & 0.813 & 0.777 & 0.772 & 0.750  & 0.800    & 0.699 & 0.700   \\
                          & \textbf{0.1} & 0.782 & 0.778 & 0.800   & 0.721 & 0.717 & 0.705 & 0.774  & 0.686 & 0.691 \\
                          & \textbf{0.3} & 0.678 & 0.673 & 0.658 & 0.616 & 0.608 & 0.631 & --     & --    & --    \\
                          \midrule
\multirow{3}{*}{ \rotcell{\scriptsize \cseer}} 
& \textbf{0}   & 0.880  & 0.887 & 0.880  & 0.87  & 0.859 & 0.846 & 0.810   & 0.791 & 0.795 \\
                          & \textbf{0.1} & 0.846 & 0.830  & 0.852 & 0.691 & 0.683 & 0.700   & 0.778  & 0.766 & 0.758 \\
                          & \textbf{0.3} & 0.747 & 0.731 & 0.735 & 0.620  & 0.606 & 0.617 & --     & --    & --    \\
                          \midrule
\multirow{3}{*}{ \rotcell{\scriptsize \pubmed}}   
& \textbf{0}   & 0.642 & 0.633 & 0.681 & 0.592 & 0.58  & 0.66  & 0.533  & 0.521 & 0.568 \\
                          & \textbf{0.1} & 0.564 & 0.548 & 0.653 & 0.570  & 0.554 & 0.556 & 0.519  & 0.509 & 0.537 \\
                          & \textbf{0.3} & 0.523 & 0.508 & 0.600   & 0.347 & 0.503 & 0.542 & --     & --    & --    \\
\midrule
\multirow{3}{*}{ \rotcell{\scriptsize \flickr}}   
& \textbf{0}   & 0.810  & 0.775 & 0.820  & 0.568 & 0.665 & 0.745 & 0.302  & 0.500   & 0.500   \\
                          & \textbf{0.1} & 0.685 & 0.541 & 0.785 & 0.541 & 0.514 & 0.680  & --     & --    & --    \\
                          & \textbf{0.3} & 0.571 & 0.505 & 0.655 & 0.484 & 0.645 & 0.649 & --     & --    & --    \\
                          \midrule
\multirow{3}{*}{ \rotcell{\scriptsize \reddit}}   
& \textbf{0}   & 0.747 & 0.825 & 0.830  & 0.730  & 0.801 & 0.814 & 0.176 & 0.500   & 0.500   \\
                          & \textbf{0.1} & 0.574 & 0.533 & 0.655 & 0.521 & 0.524 & 0.620  & --     & --    & --    \\
                          & \textbf{0.3} & 0.527 & 0.518 & 0.632 & 0.227 & 0.384 & 0.500   & --     & --    & --      \\
                          \bottomrule
\end{tabular}
\end{table}

\begin{figure}[t!]
\centering
\begin{subfigure}[b]{0.5\linewidth}
\centering
\includegraphics[width=1\linewidth]{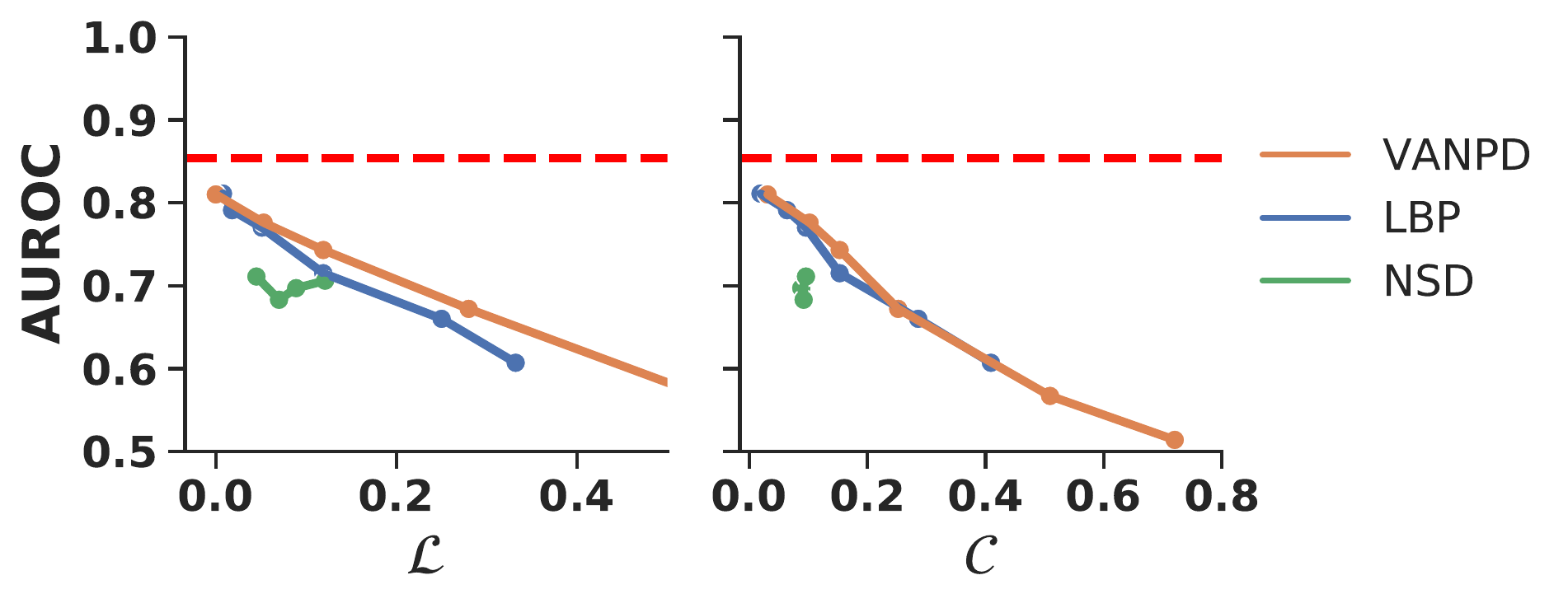}
\caption{\cora}
\label{sfig:lossconfcora}
\end{subfigure}%
\begin{subfigure}[b]{0.5\linewidth}
\centering
\includegraphics[width=1\linewidth]{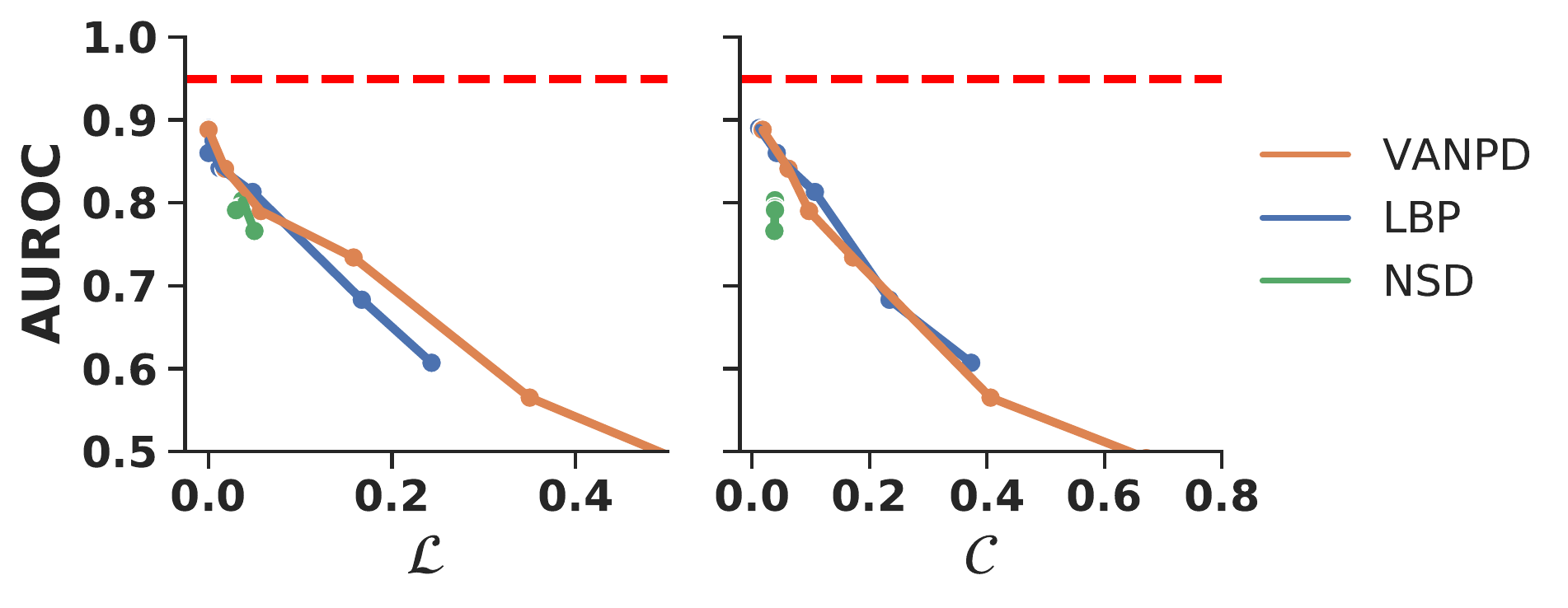}
\caption{\cseer}
\label{sfig:lossconfcseer}
\end{subfigure}%

\begin{subfigure}[b]{0.5\linewidth}
\centering
\includegraphics[width=1\linewidth]{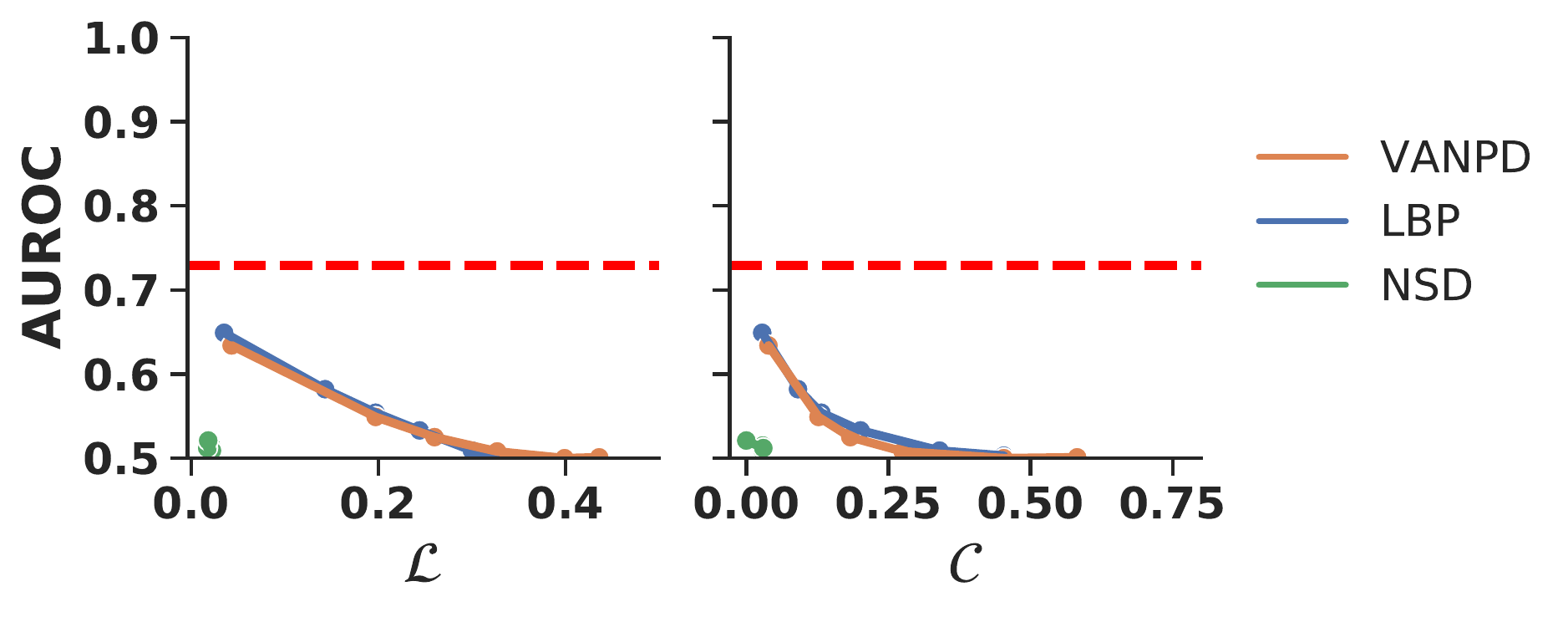}
\caption{\pubmed}
\label{sfig:lossconfpubmed}
\end{subfigure}%
\begin{subfigure}[b]{0.5\linewidth}
\centering
\includegraphics[width=1\linewidth]{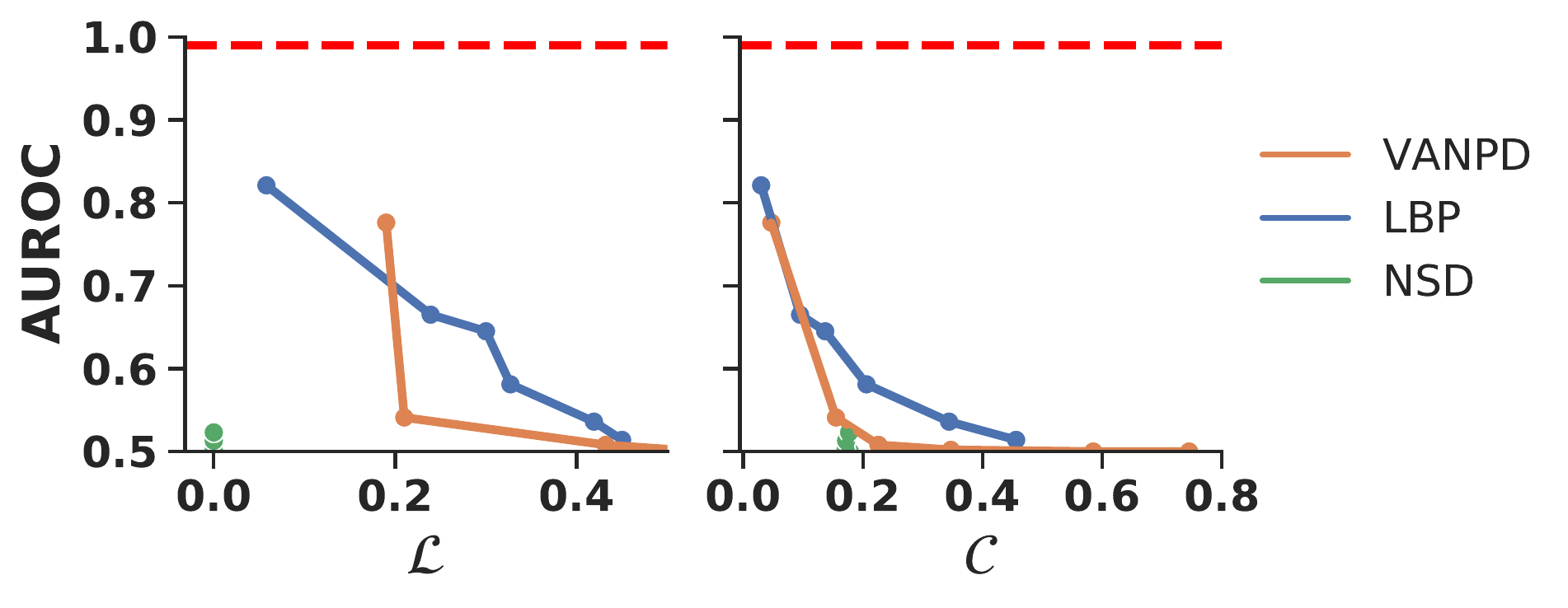}
\caption{\flickr}
\label{sfig:lossconfflickr}
\end{subfigure}%

\begin{subfigure}[b]{0.5\linewidth}
\centering
\includegraphics[width=1\linewidth]{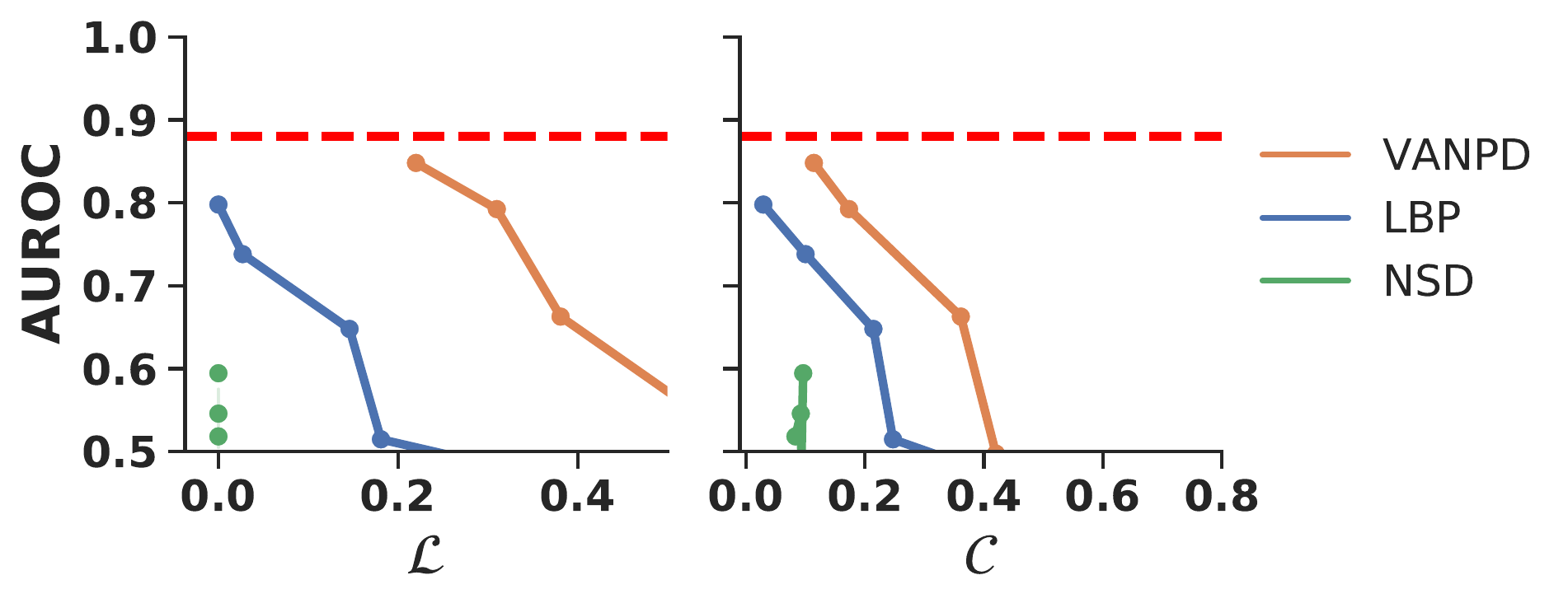}
\caption{\reddit}
\label{sfig:lossconfreddit}
\end{subfigure}%

\caption{ Comparison of defense mechanisms using label loss ($\mathcal{L}$) and confidence score distortion ($\mathcal{C}$) metric. The red dashed line corresponds to the attack AUROC when no defense mechanism was applied. We observe a similar trend for Precision and Recall. 
Table \ref{tab:labellossattackperf} shows the performance on all metric at specific $\mathcal{L}$.
}
\label{fig:labellossconfidistor}
\end{figure}

In Figure \ref{fig:labellossconfidistor}, we plot the attack AUROC (after the defense is applied) together with label loss and confidence distortion. In the following, we analyze the results for different datasets when the three different defense mechanisms were applied. Table \ref{tab:labellossattackperf} further provides the attack precision, recall and AUROC scores (after defense mechanism has been applied) and the corresponding label loss. All results corresponds to attacks in TSTF setting except for \reddit in the TSTS setting.
\paragraph{\cora} Recall that \cora is a sparse graph (with average degree 3.89 in TSTF setting) with $7$ classes. Because of high sparsity, we do not benefit much by \nsd defense which perturbs the input neighborhood of query node. Nevertheless, it achieves a drop of $15\%$ in attack's performance with a negligible label loss and confidence distortion (see Figure \ref{sfig:lossconfcora}). \lbp and \vanpd which directly perturbs the posteriors, achieves larger drops in attack performance though at the expense of higher label loss and confidence distortion. Nevertheless, for the same label loss, \lbp achieves a better drop in attack performance. For instance, \lbp achieves a maximum drop of $24\%$ at label loss of $0.3$ whereas for \vanpd, the drop in attack precision is only $17\%$ at the same label loss. At $0.1$ label loss, \lbp achieves a $12\%$ (thrice the percentage drop in attacker's inference than \vanpd and \nsd at $0.1$ label loss).

\paragraph{\cseer} For the \cseer dataset (Figure \ref{sfig:lossconfcseer}), 
the drop in attacker's inference for \lbp at label loss of $0.3$ and $0.1$ is 30\% and 22\% respectively which is two times (16\%) and four times (5\%) better than \vanpd at the same label loss. At $0.1$ label loss, \nsd achieves $12\%$ drop in attacker's performance. However, at a $0$ label loss, \nsd still achieves $9\%$ drop in attacker's performance while \vanpd and \lbp only have $1\%$ and $2\%$ drops respectively.

\paragraph{\pubmed} On the \pubmed dataset, we observe a further reduction in the attacker's inference with \vanpd achieving $25\%$ and \lbp achieving $50\%$ at label loss of $0.3$. \nsd does not incur any label loss above $0.1$. Hence, at a lower label loss of $0.1$, \vanpd and \lbp perform similarly. One possible explanation is that \pubmed only has three classes. Therefore, the maximum number of bins is restricted to $2$ which does not lead to any specific advantage for \lbp as compared to \vanpd. On the contrary, \nsd performs well with a $26\%$ and $24\%$ drop in attacker's inference at a label loss of $0.1$ and $0$ respectively making the attacker's performance largely incorrectly classifies member nodes as non-members. (Figure \ref{sfig:lossconfpubmed}).

\paragraph{\flickr} As in the previous analysis on other datasets, \lbp outperforms \vanpd by about over $15\%$ at label loss of $0.3$. It is notable that \nsd that samples the neighborhood of the query performs significantly better than \lbp with a drop in inference performance of $65\%$ at a perfect label loss of $0$. This significant drop explains the intriguing observation that the predictions of a node follow that of its neighbors (Section \ref{sec:instanceconn}). Therefore, when the neighbors of a node are distorted by the query sampling mechanism, the posterior is equally affected, causing the attack model to misclassify member and non-member nodes. 

\paragraph{\reddit} Similar to \flickr, at label loss $0$, we observe an $80\%$ drop  for \nsd,  $17\%$ for \lbp, and $15\%$ drop for \vanpd. We note that a similar drop in attacker's inference that \lbp will achieve at $0.3$ label loss, \nsd will achieve the same drop at a perfect $0$ label loss (observed at $k=2$). The observations also follow that of \flickr because of the high node degree of the \reddit dataset.

\mpara{Summary.}
We observe that \vanpd leads to a  degradation in the test performance of the target model as well as the attack performance. Although this significantly defends against MI attack, it is at the expense of the test accuracy of the target model. Binning as in $\lbp$ provides a viable strategy to reduce the amount of added noise without compromising defense. We observe that our \lbp defense is well suited for graphs with a low degree. For \lbp, setting $\psi=2$ led to a good balance between privacy and limiting label loss. We remark that both \lbp and \vanpd are evaluated on the same noise scale. 

On all datasets, \nsd achieves the lowest label loss. We attribute the observation of different defenses to the degree of each graph. Specifically, \cora, \cseer, and \pubmed have low degrees, therefore, the \nsd does not significantly reduce the attacker's inference. However, on large datasets such as \flickr and \reddit which have higher degrees, the attacker's inference reduces to a random guess (with AUC score of 0.5) with a perfect $0$-label loss. With respect to the choice of $k$, we observed that the smaller the value of $k$, the better the defense. For the current datasets, we observed that for $k>3$, there was not much degradation in attack performance.

\mpara{Comparison based on confidence score distortion.} \\
The lower the $\mathcal{C}$, the more difficult it is for an attacker to detect whether the model has undergone any defense. Moreover, a lower confidence distortion is required for applications where the target's model output posterior is used rather than just the predicted class. 
As shown in Figure \ref{fig:labellossconfidistor}, \vanpd leads to very high confidence distortion as compared to other defenses. For instance, \vanpd, \lbp, and \nsd achieves $0.70$, $0.40$ and $0.10$ confidence score distortion respectively on \cora dataset corresponding to the reduction in attack precision by $15\%$, $15\%$ and $12\%$. On the larger dataset \reddit, \vanpd, \lbp, and \nsd achieves $0.82$, $0.60$ and $0.05$ confidence score distortion corresponding to $30\%$, $30\%$ and $63\%$ reduction in attack precision.
Our result shows that \nsd achieves the lowest confidence score distortion leading to an oblivious defense and the preservation of target model's utility. \\

\section{Conclusion}
We compare the vulnerability of GNN models to membership inference attacks. We further show that the observed differences in vulnerability is caused by differences in various model and dataset properties. We show that the simplest binary classifier-based attack model already suffices to launch an attack on GNN models even if they generalize well. We carried out experiments on five popular datasets in two realistic settings. To prevent MI attacks on GNN, we propose two effective defenses based on output perturbation and query neighborhood sampling that significantly decrease the attacker's inference without substantially compromising the target model's performance.

\mpara{Acknowledgements.}
This work is in part funded by the Lower Saxony Ministry of Science and Culture under grant number ZN3491 within the Lower Saxony "Vorab" of the Volkswagen Foundation and supported by the Center for Digital Innovations (ZDIN), and the Federal Ministry of Education and Research (BMBF) under LeibnizKILabor (grant number 01DD20003).

\bibliographystyle{abbrv}
\bibliography{main}

\end{document}